# DREAMT - Embodied Motivational Conversational Storytelling

**David M W Powers**
*College of Science and Engineering*
*Flinders University*

David.Powers@flinders.edu.au

Storytelling is fundamental to language, including culture, conversation and communication in their broadest senses. It thus emerges as an essential component of intelligent systems, including systems where natural language is not a primary focus or where we do not usually think of a story being involved.

In this paper we explore the emergence of storytelling as a requirement in embodied conversational agents, including its role in educational and health interventions, as well as in a general-purpose computer interface for people with disabilities or other constraints that prevent the use of traditional keyboard and speech interfaces.

We further present a characterization of storytelling as an inventive fleshing out of detail according to a particular personal perspective, and propose the DREAMT model to focus attention on the different layers that need to be present in a character-driven storytelling system.

## 1 Introduction

Embodied Conversational Agents (ECAs) have moved from being research artefacts a decade ago to an integral part of a modern personal computer or mobile phone interface and expectation. However, apart from their speech capabilities, their language technology has moved on very little since the early conversational agents of the 1960s and early 70s, notably ELIZA (Weizenbaum, 1966) and PARRY (Colby et al. 1972). The ELIZA 'doctor' and the PARRY 'patient' themselves reflect two sides of the coin we examine in this paper, which focuses on the *roles* played by agents. The AIML-based ALICE system (Wallace, 1995/2008) and successful conversational/Q&A bots in general (Shah et al., 2016) still rely largely on templates, with AIML providing an open Artificial Intelligence Markup Language that facilitates writing Talking Head ECAs and toolboxes (Luerssen et al., 2010).

We follow in this tradition while downplaying the syntax in favour of semantics – it is very easy to program a sophisticated grammar and use it for generation, and a generative approach to sentence understanding and disambiguation within this probablistic controlled grammar is thus also straightforward. Although much of our work has focused on learning and disambiguating phonology, morphology, grammar and semantics (Powers, 1984,1997ab; Huang and Powers, 2001,2003; Yang and Powers, 2008), we eschew 'black box' statistical or connectionist learning and embedding techniques (which are strongly biased by web data and severely overfitted by deep learning). Rather we concentrate on allowing several different personalities or characterizations to produce different stories or dialogues from the same underlying ontological representation.

Whereas it is straightforward to recognize a story in other forms than the written or oral, including plays, movies and multimedia (Ryan, 2007) or even computer games (Riedl, et. al, 2011), storytelling has a role that is perhaps not so obvious in everyday conversation, and in interactions with students or patients, but we argue that thinking in terms of storytelling offers the opportunity to improve the role of AI in such ECA interventions – both in terms of conveying content in a way that is in some sense optimal, and in terms of conversing in a human like way. Moreover, evaluation of an ECA is normally two-pronged, one focus being the subjective naturalness, friendliness and acceptability of the ECA interface and dialogue, and the other being the objective effectiveness of the intervention (Powers, et al. 2008, Milne et al. 2010,2011,2018; Stevens et al. 2016).

A related theme building on our multimodal research (e.g. Lewis and Powers, 2002; Atyabi et al. 2012; Fitzgibbon et al. 2015; Grummett et al.

2015) is the development of Unconscious AudioVisual and Brain Computer Interfaces (UCI/ABCI), with a particular focus on educational and health interventions, including an audiovisual brain controlled wheelchair and a multimodal authoring system for people with disabilities – in both cases these are high-level model-driven systems where the computer is aiming to predict behavior and intention as much as possible, avoiding step-by-step and character-by-character control regimes. This involves understanding the user's character and situation and building an appropriate story.

Understanding how to tell a story or direct a scene is essential to making ECAs and UCIs work whether explicitly using language or implicitly directing behaviour.

## 2 Grounding and Characterization

Although AI researchers do not normally think of themselves as trying to pass the Turing Test or evidence a mind (Powers, 1998), failure to meet expectations leads to a jarring experience, and Mori's (1970) uncanny valley. The Turing Test (TT) was conceived by Turing (1950) as a 'pen pal' game, but Turing felt that a successful winner would need to have robotic sensors and capabilities, and this was captured in Harnad's (1992) 'Total Turing Test (TTT)' and (1990) 'Symbol Grounding Problem', contending that a TT would need real world ontology that interpreted sensory perception of the real world, unconvinced by a tradition of simulated robot worlds that goes back to Block (1968) and Winograd (1973) and Hume (1984).

Block (1968) emphasized the mantra that 'the verbs are the parts' of the object and used a lattice-like representation, Winograd (1973) emphasized a LISPish 'procedural' representation, and Powers (1984) used a tree-like Prolog 'space-time-state' representation. Such representations were argued to be more bioplausible than the Chomskian and Fodorian claims that different 'innate' representations (or modules) were required for different aspects (or modalities) of language. Whereas earlier work had talked about 'semantics' and translated into ungrounded representations, Powers (1984) switched the focus from 'semantics' to 'ontology' (meaning 'understanding of what is'), to avoid this misuse of 'semantics' without 'grounding', defining a 3D simulated robot world model involving fixed, mobile and motile objects suitable for concept learning as well as language learning (Hume, 1984; Sammut 1985).

Steels (1999) met Harnad's (1990) criterion of real world grounding a real world paradigm of language learning and development with physical robots which also fit this paradigm, with (basic) language being emergent from the robots' actual sensory-motor representations, and over the last 20 years we and many other researchers have further explored both real world and toy world grounding with robotic agents and other embedded devices (from the humble photocopier to a futuristic spaceship). It should also be observed that mixed modality grounding is possible through the web and its stored information, conversations, and multimedia, including live video feeds and virtual agents (including ECAs). This may be seen as a sensory-motor surrogate, but mostly lacks the important motor dimension of true grounding, including the sensory/motor and drive/affect aspects of a child's interaction with the world that are part and parcel of it learning not only syntax and semantics, but a rich social and cultural understanding of our environment.

The 'grounding' factor in 'real understanding' was also the basis for Loebner's requirement of a sensory-motor component for his Gold Medal Loebner Prize, working with Powers (1998) to realize this with a show-and-tell scenario that involved introducing a favorite doll or other toy and telling a story about their relationship with it.

At present, the focus for demonstrating humanness and personality in ECAs remains largely a matter of providing it a human-like family and work/hobby backstory, along with clever emulation of superficial features (Shah, 2011ab; Shah et al., 2016), including in modern ECAs providing a range of prerecorded voices and emotional registers, as well as programming expressive faces and light-hearted personality-reflecting responses. Such techniques (even if canned) can have an impact both on user acceptance and on the success of an educational or health-related intervention (Stevens et al., 2010).

The flip side of this is to develop an interlocutor model. In a story dialogue, the computer scripts all characters, but in a conversational agent, it is helpful to model and predict human responses. This ability to model and customize to a particular human user has again been extensively explored in educational and health contexts (Wittwer et al., 2010; Jones et al., 2016), where selection of an appropriate intervention/treatment template for the student/patient can be viewed as a diagnostic step.

For the purposes of the present model, we accept that it is convenient to understand both language and ontology in terms of a common phrase structure model with tree representations and argument attributes, both of which are managed in Prolog's DCG formalism, which can also handle traditional generalizations above and below word-to-sentence level grammar, such as Pike's Tagmemic Grammar (Pike 1943, 1967; Pike and Pike 1982) as well as subsuming both taxonomic and meronymic relationships in a general logical spatiotemporal representation. Pike's broad use of the Tagmemic formalism also suggests how we can extend this common representation principle to cover the modeling and representation of character and reader through the use of dynamic character attributes.

We follow Hayes (1979), Powers (1984) and Hume (1984) in structuring our universe into a sequence of time-and-place-stamped events under the constraint of simulated physical laws. Events in turn have object models with all the information needed for 3D rendering, as well as all the attributes necessary for understanding them as characters (distinguishing motile characters that can move themselves, which operate under robotic control, from mobile characters that can be pushed around or grabbed and lifted, to fixed objects that can't). Some of the first stories in this world concerned a dog chasing a postman or a cat chasing a mouse, over the years developing from crude line drawings to sophisticated 3D renderings.

Initially, our focus was learning and then using nouns, verbs, adjectives and basic grammar, with the learner being human controlled and the other characters being directly programmed (with simple randomly-initiated motivation-driven behaviours). Other systems have been developed and evaluated for specific educational or health applications. In this paper our focus is on learning and modeling more sophisticated motivations and behaviours, for human-level characters who have both a sense of identity and an understanding of other characters.

## 3 Fiction, Fantasy, Fairytale and Folktale

There are some domains where the human and the computer are on a more equal footing, in particular in the description of imagined worlds in fairytales, and science fiction/fantasy. Here the reader/hearer is reliant on the storytelling prowess of the author to convey the critical information about worlds with varying degrees of relation to our real world, including conveying abstract concepts (like heroism) in a simplified context (like fairyland) in which many of the restrictions and limitations of the real world are removed (like magic). This is thus the domain of our present storytelling focus.

Storytelling is somewhat different from general conversation in several respects that are pertinent from a conversational, motivational and educational perspective:

1. We *operate* in an invented world (story) only *similar* to our real world (history);
2. We *plot* a storyline, as well as details of the *world* and its *population* and *artefacts*;
3. We *write* from some *perspective* (narrator or specific point-of-view characters);
4. We *select* what to write/talk about relative to the *nature* of both *narrator* and *audience*;
5. We *obey* strictures and conventions that relate to the *genre* more than *subject*;
6. We *speak* both/all sides of any dialog (respecting the *nature* of the character);
7. We *present* information in a (logical or chronological) order that fits our genre.

By focussing on storytelling in a fantastic world, we gain the same kind of advantage as we have with toy worlds and toy robots. The obvious advantage is simplification, but this hides a more fundamental advantage: we avoid much of the problem of selection and abstraction.

When we write an (auto)biographical story, or answer a simple 'what did you do in your holidays?' question, we have a huge amount of information available to us, and what we need to present is a function of the author's goals and self-perception as well the reader's background, needs and expectations. As we move back in time, history becomes more story as the 'natural selection' by the events and people of preceding generations abstracts out details and leaves the storyteller to concoct myths around the bare threads, reinventing the detail that will provide an engaging yarn and a culturally acceptable moral targeted to an ecologically appropriate theory of mind.

At another level, every story needs to be a mystery – the clues need to be laid out, and interest needs to be maintained through engagement with characters at risk or heroes on a quest (or an investigation), and an outcome in question. Thus although mysteries and thrillers are separate genres, good stories will have elements of mystery and elements of thrill. Ideally, we should maintain the suspense of who done it, or who will win, and how they did it or will do it – even when we know the story, or we understand that the hero will come through in the end.

## 4 Information Retrieval/Literature Review

Something similar occurs in Information Retrieval and Summarization. We have in general different opinions and claims, as well as different levels of knowledge and ignorance reflected in the 'facts' that we are aiming to summarize. The classical example here is the history of an invention, where different people have valid claims, and reinvention of the wheel is a necessary consequence of poor communication, incomplete information, and active or passive filtering (e.g. by journals or professional societies, or by religious or political organizations). Writing a Literature Review ideally recovers these 'origins' threads, but in practice puts a spin on the history of the field and can specifically seek to justify the approach taken in a specific project.

This is currently the most well-developed application for our storytelling approach, but is nonetheless still at a very early stage. One such project was to provide a novel computer interface that allowed users with disabilities to communicate without the ability to speak or type. Rather than using a character-by-character approach we moved to a higher level approach informed by the user's actual experience – one aspect of this project is to automatically or semi-automatically develop a 'life reel' that selects key events that might be the basis for a story as part of either a conversation or a written message (our focus here includes elderly people at the onset of dementia). This illustrates that the modality of storytelling is not just limited to text or speech – here the initial story is composed as an edited movie.

Another aspect of this disability-oriented project is to be able to move around and act on the world in a natural way, including guiding a wheelchair around a multistorey building and the local public transport system, while avoiding the cursor-like or joystick-simulating left/right forward/back control of the traditional brain computer interface. This builds on our earlier work with robot navigation (and in fact the wheelchair shares a mid-level interface with several families of robots). Moving our characters (self-moving 'motile' objects) around an environment, pushing, grabbing and/or carrying other ('mobile') objects, and navigating the built infrastructure and obstacles ('fixed' objects) is the same kind of storytelling we used originally in our robot worlds for learning and teaching natural language – again illustrating storytelling through a non-linguistic modality.

Recently, our research personnel and students have begun to see that this kind of interface would be useful for able-bodied people too – we are all effectively disabled in contexts where we can't use our hands (e.g. we are holding something) or we can't speak (e.g. we are in a library). However, the storytelling tools we are building also provide useful tools for anyone who is trying to write a report or navigate a building or a city.

Our model here is that we provide an exploration interface that actually stores information in a way that allows it to make predictions. In the case of the information retrieval and literature review application, it can group together and organize related work, and then the next step is to turn that into a component of the story that is being told, or the path that is currently being taken. A high level selection approach is mostly all that is needed to achieve this, although the ability to drop down to a lower level for more control is also provided – however at the moment this capability is very crude and clunky compared to the ease of operation with high level selection.

We conclude this section with a summary of the kinds of applications we have been building where storytelling seems to have an important role, forcing a more comprehensive ontological representation and more realistic semantics than the usual template-based or statistical/neural system.

1. Language/phonology teaching/learning;
2. Social skills training for kids with autism;
3. Memory assist/training for elderly people;
4. Wheelchair for people with quadraplegia;
5. Computer GUI for those with quadraplegia;
6. Motivational interviewing in health;
7. Multirole ECAs: teacher/student/peer/...

The issue of role is critical in storytelling as well as in traditional linguistics (providing a semantic counterpart to the grammatical slot). For a traditional ECA, the ECA plays a single role, but in our educational and health training applications, we have found it useful to be able to play multiple roles (usually a teacher or professional role plus a student, patient or client role, but often a peer role is helpful to illustrate good and bad practice without the direct imprimatur of the teacher). Thus it is necessary to outline a story framework that our subject fits into, rather than providing a rigid conversational template with canned questions and answers.

Currently we are also exploring the storytelling approach directly with (science fiction/fantasy) authors and stories. Here the idea is to explore the tools that authors find useful, and the limitations that they find frustrating, to allow developing tools that can in the short term assist authors with disabilities, and in the longer term extend the tools to augment the productivity of any author.

## 5   Storytelling by layer – DREAMT

We now introduce the DREAMT mnemonic and the modality independent levels of storytelling and that we feel are important based on the research we've discussed. Most if not all aspects of the DREAMT model have arisen from or been explored in some aspect of our implemented research systems, but currently only at a primitive and relatively unintegrated level. However, this experience leads us to formalize and elaborate the DREAMT model mnemonically as follows:

- Description/Dialogue/Definition/Denotation
- Realization/Representation/Role
- Explanation/Education/Entertainment
- Actualization/Activation
- Motivation/Modelling
- Topicalization/Transformation

### 5.1   Dialogue vs Descriptive Detail

The primary function of storytelling is arguably to convey information, although secondary goals relating to the effect on the audience will often be dominant. For all these goals it is essential to reach the audience at an appropriate level, including establishing a shared basis for understanding the story.

A primary choice we must make in story telling is whether we are going to 'show' or 'tell', and this relates also to the question of background vs foreground information as well as to Plato's diegetic vs mimetic distinction. Modern novels tend, like plays and screenplays, to be dialogue heavy, and with dialogue we may also be tempted to lump in monologue and introspection (as well as science fantasy tropes such as telepathy). There is however a distinction between conveying information to the reader as narrator, providing putative introspection or retrospection, and reporting a conversation conveying the same content to another character – the modern novel or screenplay tends to avoid the traditional conventions of the stage, and aims to ensure that monologue and dialogue respect rather than define the nature of the character.

Modern fiction however tends to be detail heavy, like advertisements seeking to provide authenticity by the artifice of providing detail such as the brand of products used. Making this choice of 'device', and using dialogue or description effectively, depends on decisions and intentions that relate to other levels of our storytelling model. That is the decision should not be arbitrary but must have a specific purpose (which is usually to provide useful information without unnecessary overload, although a detective story or thriller may reverse this by providing misleading clues, in descriptions, and false information, in dialogues).

The precise terminology used also figures here, and again the explication with overt definition or implicit denotation depends on other levels of the storytelling model, and in particular a developing model of the reader's and hero's understanding of and deictic interaction with the world. Note that for dialogue in a novel, these are inherently different perspectives but commonality is required for effective deixis. Sometimes the difference in reader/character knowledge is itself a device – often suspense is derived from the mismatch artifice of allowing the reader to know what the heroine does not, or vice-versa (e.g. a plan or clue or interpretation is withheld from the reader).

An intermediate level between description and dialogue comes with the modern idea of a point of view that can shift between characters and show what they are thinking, rather than the traditional idea of a narrator. Here there is also a stylistic (or genre-related) decision to be made as to whether first or third person will be used, and some writers also use the second person (or inclusive first person) to bring you into the scene or to think through a situation with you (that is there is an implicit dialogue with the reader, or with a generic person, as we/you/one might say).

Providing a character point of view (PoV) gives the author an easy way to put us inside the characters head and create empathy and investment. But character is not just about what one thinks or says

– it needs to be demonstrated in practice by what our characters do and what they risk.

## 5.2 Realization vs Representation

Once we have decided the foreground/background show/tell mode for relaying a feature of the story as represented in (we assume) some kind of tree-like or lattice-like representation, this naturally needs to be sequenced into words, phrases and sentences. This is arguably the main focus of research in any form of Natural Language Generation (NLG: Reiter and Dale, 2000; Gatt and Krahmer, 2018), but one that we will neglect here to focus on the other levels of the story telling model that influence the form it takes.

In our work, the transformation to and from an underlying representation is ideally learned (with a minimum of supervision) rather than programmed, and itself covers multiple levels, including the syntax level (word to clause or sentence), the morphological level (phoneme or grapheme to word), and the phonological level (sound to phone and phoneme). Our focus here is not the mechanical task of translating from a tree representation to a sentence template, but the higher level task of determining the modal form of sentences and their cohesion into higher level units, using a simplified tagmeme-like model similar to that of Pike and Pike (1982) that has been used extensively both within and beyond traditional linguistic analysis, originating with Ken Pike's invention of the phoneme (1943, 1947) and generalizing to stories, discourse and human behaviour in general (1964,1967,1973,1982,1986).

Tagmemic grammar characterizes the traditional non-terminals and rewrite rules of phrase structure grammar (PSG) in terms of slot and (filler) class, augmenting them with attributes denoted role and cohesion that fit well within an attribute grammar or unification grammar framework.

Powers (1984,1989) characterizes the slot as an intrahierarchical relationship, as captured also in categorial grammar – the slot name is more specific than a traditional part of speech in that it carries the power of a categorial fraction and new slots can be learned dynamically. Role is categorized as interhierarchical in the sense of relating a semantic role to a syntactic slot, and is the main tagmemic cell in focus here – the role name specifies the specific function of the slot and constrains the filler class, conveying things like voice and mood and propagating constraints (e.g. subject slot has undergoer role in passive voice). The fourth component of the tagmeme, is transhierarchical, carrying constraints from one part of the parse tree to another (e.g. gender, person and number as linguistic constructs rather than ontological realizations, tables being feminine, crowds being singular, etc.)

For the purposes of the present model we think more in terms of layers than independent hierarchies, and the tagmemic roles serve to pass constraints between layers, enforcing constraints within the realizing modalities. In practice, in our Prolog-like implementations, the four cells of the tagmeme are just four independently specifiable attributes of a Definite Clause Grammar (DCG).

## 5.3 Explanation and Education

We now move to the first level that is *not* commonly included as part of an ECA or NLG system, but should be an essential part of any story telling system, including those without a specific educational function, and especially those that purport to be an assistive technology.

In our educational and motivational systems, the computer normally plays the teacher role and the human user the student role, but for teaching a pedagogic or motivational approach we can reverse this to exemplify different strategies and give the user the feel of being on the receiving end. Similarly, we can also allow for specific teaching of a Natural Language Learning system (NLL) by a trusted user, and this has been the focus of much of our research (in a hybrid supervision paradigm).

Explanation is also well known in the context of both Expert Systems and Machine Learning. In an Expert System, an expert may ask why a question is being asked, or a user may ask how a conclusion has been reached. In Explanation-Based Learning (EBL), examples may be used to help make a Natural Language system more efficient or efficacious in a particular context, or to help a Natural Language system interpret instructions with the help of a worked example (Samuelsson and Rayner, 1991; Delisle et al., 2005).

In the context of storytelling, education and explanation are major sources of direction for the storyteller, although an emotional explanation may have more weight than, or even mitigate against, the logical and justifiable course of action. The storyteller needs to explain how situations come about and why actions are taken, and as emphasized earlier ensure that the framework is in place both to understand the information being conveyed and to disambiguate the language being used.

This relates also to the major decision of whether to be logical or chronological, and when to bring in past information as background for understanding future events. There is also the related question as to how much the reader is allowed to hang in confusion or suspense before an explanation is given.

The logical ordering normally has a pedagogical goal, aiming to prepare the reader in advance for the complex new concepts and events that are the primary focus. The chronological ordering aims to engage the reader and take them forward chronologically in at least a topologically local sense (that is focussing only on the place where we find the principal character we are following). In either case, it can be frustrating not knowing why spatially or temporally distant, and apparently unrelated, events and people are being presented, how they will come together, and what investment is necessary in terms of remembering all this information – this needs to be part of a reader model.

An alternate approach is to pursue key characters chronologically through key events, and fill in backstory by flashback when it becomes relevant. This can also be confusing or annoying to the reader, and we again have a decision to make about whether and how this is marked for the reader (as introspective reflection or dialogue, etc.) The storyteller needs to make things *plain* to the reader (ex*plan*ation) and *lead* the reader to an understanding of the course of events (e*duc*ation).

The same also applies to the characters. Readers need not only to understand things for themselves, but to understand what the characters know and how they have come to take their particular paths through the story.

However, explanation and education should *not* necessarily be undertaken explicitly. It is often better to show how things work out based on the characterizations of the universe and its inhabitants. Brief explanations of laws (by character or narrator) can prospectively or retrospectively label a causal sequence that is played out actively on stage – this is the heart of explanation based learning, where the examples come first and the rules are induced, or the theory is provided first and the examples explicate how they actually apply in practice.

Closely related to this is the question of planning and execution. Although logically the plan comes first, it is often better for the story to summarize this as "I've got an idea…" However, "no plan survives contact with the enemy", and sometimes revealing the plan is necessary to provide the backdrop for what goes wrong (Delisle et al., 2005).

## 5.4 Motivation and Activation

Activation and actualization result from putting your characters in motion or understanding their motivation, and in movies, multimedia or ECAs this mimetic component will be conveyed multimodally rather than descriptively. The concepts of motivation and activation are traditionally juxtaposed and should be understood together, as motivation and activation dynamically influence each other through feedback paths in both the character and the reader.

For this purpose, we regard sophisticated props (from spaceships to computers, from fairy tale castles to everyday pets) as being characters too. The simple programming of a computer, or the ordinary drives of a cat, are in essence not much different in character from the functional behaviour or the consistent behaviour of a character who is being true to herself. Of course, this well-behaved law-abiding behaviour can be disrupted by shifting the character into a new situation – and typically this happens in the first act of a traditional three act structure.

The hero or heroine is thrown into deep water, sink or swim. Their existing programming is not enough to cope, and the strengths and weaknesses of their characters are brought into focus. They need to step up to the challenge, and usually also need to shift their motivational focus and drive from their own interests and safety to a bigger picture. The characters can only play hero/heroine when someone else is at risk, usually a lot of someones. They are moved out of their comfort zone both physically (motion) and psychologically (emotion). The effect on them at the emotional level (affect) leads them to take action (activation) and realizes possibilities (often negative possibilities, dangers, risks and fears, but usually also positive possibilities, opportunities and the potential for making a mark on the world)

The representation of the storyline and our logical path is actually fleshed out and driven by the nature and motivation of our characters as represented in models of themselves and others, where models of others typically start out as being self-like ("the charitable assumption") but then get modified as we find out more (an alien, a child, a professor, a student, a psychologist, an engineer). What is perhaps less obvious is that there will also be, implicitly or explicitly, models of the target audience. In the case of a computer game or ECA intervention, this is likely to be an active participant rather than a mere reader.

Part of the challenge of a storytelling system is thus to take our models of the characters and participants and allow them to interact, each learning the characters of the other characters. This ideally involves understanding them in a grounded sense, but the more different from 'us' they are the less common ground we have, and the less well-grounded our model of them will start out. Nonetheless our shared sensory-motor interaction and experiences will build up this foundation, and will change the internal activation (of our character or character model) that is triggered by the new situation, characters and events. This in turn will lead to actualization along the path of our story.

Hopefully… if we have managed to capture appropriate motivation and models for our characters, consistent with our plots. But sometimes characters, and stories, take on a life of their own!

## 5.5 Topicalization and Transformation

Up to this point, our focus has been on how the story is going to unfold, and the high level decisions and low level character models that influence this. But so far, we still have tree-like representations tagged with spacetime information. So we now return to the question of translating these ontological representations into actual words, phrases, clauses, sentences, paragraphs and stories.

Although we don't actually transform linguistic structures from one form to another, it is tempting to think in transformational terms. We should also note that dealing with complex sentences is still a challenge for current NLP systems (where even "gold standard" tagged treebanks and corpora have many errors due to employing taggers and parsers that really don't understand complex sentence structures, are poor at disambiguation, and use evaluation metrics that hide how bad they are).

NLG doesn't have the same level of problem here, because we can only generate correct grammatical sentences using fixed (and potentially restricted) grammars. However, this does become more complex once we introduce ideas of characterization involving register, accent, etc. Moreover, generating text using a sophisticated grammar actually offers the potential to generate artificial corpora to evaluate and improve our NLP tools. Thus unlike NLP we are not directly concerned with high levels of sentence complexity.

A key aspect to generating a sequence of clauses is to track the flow of topic and reflect this in our sentence structure using topicalization.

Topicalization brings the object or event of interest to the front of a clause, but does not necessarily make it the subject or complement of the clause. This increases complexity only slightly if passive voice is used to achieve this, but in English other forms of fronting are used only in interrogatives and imperatives and other specific constructions which do tend to add complexity:

- That is something I won't put up with.
- Where are you going to? I'm going to England.
- Where are you going? I'm heading off to lunch.
- Who is the boy you like? John is the boy I like.
- Who did you see yesterday? [I saw] John. /John is the boy I saw./It was John./I did see John.
- Who did I see at the mall yesterday? It was me [you saw at the mall yesterday].
- The boy I saw yesterday, the girl I didn't.
- The boy [whom] I saw yesterday was there again.
- The boy, whom I saw yesterday, was there again.

How much complexity is appropriate depends on the nature of both the character whose speech (or PoV) is being presented and the sophistication of the reader, as well whether the presentation is written or oral. Also, when questions are involved, there is a question as to whether the same template will be used (which involves low cognitive load) or an abbreviated answer will be given (which involves anaphora that must be resolved) or whether a crosscut answer is given (which may simply use a shorter form, or may reflect an unexpected or potentially ambiguous answer).

The first example shown illustrates what naively looks like dangling prepositions, but in fact is more correctly an inseparable verb involving particles that belong with the verb rather than with the fronted object – Winston Churchill got it wrong with his "up with which I will not put" in an attempt to follow the prescriptive rule about not ending a sentence in a preposition. The second example is similar, and perfectly correct and natural English. But that doesn't mean that a pedantic prescriptivist speaker *in the story* won't move the prepositions.

The final two examples above add a little more complexity as we have a subsidiary clause whose precise role depends on the punctuation. In the last example, the 'whom' takes the role of the (fronted) object in a clause that is merely parenthetic. Without commas it identifies which boy we are talking about in the main clause, and without the 'whom' (or 'that') the boy is the (fronted) object of the subsidiary clause.

A related problem is conjunction. This, of course, brings up the complexity and errors associated with the use of 'and' (with issues like "Me and Jim came…" vs "… on behalf of my wife and I" – both of which are formally ungrammatical but potentially appropriate to a character register). More significantly, 'and' can connect any part-of-speech, including elided clause components (e.g. "The boy hugged and the girl kissed their mother.")

However, other connectives also come into play and have associated syntactic and role constraints, including 'if', 'when', 'after', 'as', 'however'. In addition, there are complex usages of participles and infinitival clauses ("Driving past the school, I saw him hit[ting] the girl.")

At present our focus is on standard correct but idiomatic English, including anaphora, elision and fronting. However, our focus is the decisions as to which of many alternate constructions to use based on the constraints of storyline and characterization. In general, we take the perspective that there is no such thing as perfect synonymy or equivalence. Every lexical or syntactic choice conveys a different shade of meaning or provides a different focus.

Our focus on storyline and characterization, including decisions around explanation and chronology, provides us with an 'ideal' order for clausal units. Generally speaking, paragraph level order involves the first sentence introducing or picking up a new topic. As new objects (or events) are introduced, the system will in general already know whether a new object will become the focus, or the existing subject will be. If we are finishing a sentence with a focus on a specific new object, it will usually be topicalized even if the next clause involves the same actor as the preceding clause. This heuristic will also tend to avoid us taking longwinded excursions. Compare:
- John bought a hot dog. It was delicious.
- John bought a hot dog. He ate it and thought it was delicious.
- John bought a hot dog and tore into it hungrily. He'd never tasted anything so delicious before.

The first example topicalizes the previous object, while the second description redundantly includes a clause that is implicit both in the concept of 'hot dog' (you eat it) and the concept of 'delicious' (referring to something you eat). The final paragraph introduces a new concept (John is hungry) as well as a word ('taste') that strengthens the implication of 'eat' from 'delicious' (it could also smell delicious). Most importantly, it brings the focus back to the original topic (hungry John).

Note that the decision about how to join event descriptors (incipient clauses) into sentences and paragraphs is parallel to how these same descriptors can be turned into a virtual world 3D rendered sequence or a sequence of real-world robotic actions (moviemaking). Just as sentences need to be joined coherently, so goal-oriented actions need to be connected smoothly into plans and scenes, with transitions that take into account position, speed and acceleration (and potentially higher derivatives) of the objects and robot/character parts that are being moved or portrayed.

## 6  Conclusions

We have outlined a straightforward approach to storytelling with two facets. First, we avoid black box techniques and go back to basic Linguistics with a simple phrase structure grammar and a widely used approach to grammatical analysis that has been employed on hundreds of languages, and has previously been demonstrated to allow a variety of types and levels of learning in a simple (graphically extended) Prolog implementation. Although our examples have focused on English, we have proposed a general cognitive modeling approach, DREAMT, that guides the sequence of modeling and decision steps needed to tell a story based on an abstract representation of a universe, a plot and its characters. Future work will include learning and constructing such representations based on existing stories and common tropes.

DREAMT avoids committing to 'surface form language' till the very end, modeling and representing both the ontological events, objects and characters, and the desired (chrono)logical and motivation/activation sequence of descriptions that will be realized in actions, sentences or utterances.

The final translation step involves making specific decisions about the syntactic forms used to realize a sequence of sentences (in terms of text) but is also strongly motivated by our experience in generating utterances (in terms of audiovisual speech) and robotic behaviors (in terms of both simulated and physical robots, including a variety of autonomous ground, marine and aerial creatures and vehicles).

## References

Adham Atyabi, M Luerssen, SP Fitzgibbon, DMW Powers. 2012. Evolutionary feature selection and electrode reduction for EEG classification. *IEEE Congress on Evolutionary Computation (CEC)*.

H. D. Block, J. Moulton, and G. M. Robinson, 1975. "Natural Language Acquisition by a Robot", *Int. J. Man-Mach. Stud.,* Vol. 7, p.572ff.

Kenneth M. Colby. 1972. Turing-like Indistinguishability Tests for the Validation of a Computer Simulation of Paranoid Processes. *Artificial Intelligence* 3(1-3):199–221

Sylvain Delisle, Stan Matwin, Jiandong Wang and Lionel Zupan. 2005. Explanation-based learning helps acquire knowledge from natural language texts. In: Ras Z.W., Zemankova M. (eds) *Methodologies for Intelligent Systems.* ISMIS 1991. Lecture Notes in Computer Science (Lecture Notes in Artificial Intelligence), vol 542. Springer, Berlin, Heidelberg. pages 326-337
https://doi.org/10.1007/3-540-54563-8_96.

Sean P Fitzgibbon, Dylan DeLosAngeles, Trent W Lewis, David MW Powers, Emma M Whitham, John O Willoughby, Kenneth J Pope. 2015. Surface Laplacian of scalp electrical signals and independent component analysis resolve EMG contamination of electro-encephalogram. *International Journal of Psychophysiology* 97 (3), 277-284

Phyllis Jones, Catherine Wilcox, and Jodie Simon. 2016. Evidence-Based Instruction for Students with Autism Spectrum Disorder: TeachTown Basics. In Cardon, Teresa A. (ed.) *Technology and the Treatment of Children with Autism Spectrum Disorder,* Springer, pages 113-129.

Stevan Harnad, 1990. The Symbol Grounding Problem. *Physica D: Nonlinear Phenomena* 42 (1-3): 335-346

Stevan Harnad, 1992. The Turing Test is not a trick: Turing indistinguishability is a scientific criterion. *ACM SIGART Bulletin* 3 (4): 9-10.

Patrick I. Hayes, 1979. "The Naive Physics Manifesto", in D. Michie, ed., *Expert Systems in the Micro-electronics Age,* Edinburgh University Press, Edinburgh, Scotland, pages 245ff.

David Hume, 1984. *Creating Interactive Worlds with Multiple Actors*, BSc. Honours Thesis. Electrical Engineering and Computer Science, University of New South Wales, Sydney.

Jin Hu Huang, David M W Powers, 2001. Large scale experiments on correction of confused words, *Australian Computer Science Communications* 23 (1), 77-82.

Trent W Lewis, David M W Powers, 2004. Sensor fusion weighting measures in audio-visual speech recognition. 27th Australasian conference on Computer Science.

Martin H. Luerssen, Trent W. Lewis, David M.W. Powers. 2010. Head X: Customizable audiovisual synthesis for a multi-purpose virtual head. *Australasian Joint Conference on Artificial Intelligence*, pages 486-495.

Marissa Milne, MH Luerssen, TW Lewis, RE Leibbrandt, DMW Powers. 2010. *International Joint Conference on Neural Networks* (IJCNN), 1-9

Marissa Milne, M Luerssen, T Lewis, R Leibbrandt, D Powers. 2011. *Conversational Agents and Natural Language Interaction: Techniques and Effective Practices*, (pp 23-48). IGI International.

Marissa Milne, P Raghavendra, R Leibbrandt, DMW Powers. 2018. Personalisation and automation in a virtual conversation skills tutor for children with autism. *Journal on Multimodal User Interfaces* 12 (3), 257-269

Steven Minton, Jaime G. Carbonell, Craig Knoblock, Daniel R. Kuokka, Oren Etzioni and Yolanda Gil. 1989. Explanation-Based Learning: A Problem Solving Perspective. Artificial Intelligence 40(1-3):63-118.
https://doi.org/10.1016/0004-3702(89)90047-7.

Kenneth L Pike. 1943. *Phonetics, a Critical Analysis of Phonetic Theory and a Technique for the Practical Description of Sounds*. University of Michigan Press, Ann Arbor.

Kenneth L Pike. 1947. *Phonemics: a Technique for Reducing Languages to Writing*. University of Michigan Press, Ann Arbor.

Kenneth L Pike. 1964 Discourse analysis and tagmeme matrices. In Elmer Wolfenden (ed.)*, Papers on Philippines Linguistics by members of the Summer Institute of Linguistics*. Oceanic Linguistics 3(1):5-26.

Kenneth L Pike. 1967. *Language in Relation to a Unified Theory of the Structure of Human Behavior*. Mouton, The Hague.

Kenneth L Pike 1973 Science fiction as a test of axioms concerning human behavior. *Parma Eldalamberon* 1(3):6-7.

Kenneth L. Pike and Evelyn G. Pike. 1982. *Grammatical Analysis*. Summer Institute of Linguistics and University of Texas, Arlington.

Kenneth L Pike 1986 A further note on experimental clauses in discourse. In Benjamin F. Elson (ed.), *Language in global perspective: Papers in honor of the 50th anniversary of the Summer Institute of Linguistics, 1935-1985*, 135-138. Dallas: Summer Institute of Linguistics.

David M.W. Powers. 1984. "Natural Language the Natural Way", *Computer Compacts* Vol. 2 (3-4): 100-109.

David M. W. Powers and Christopher C. R. Turk. 1989. *Machine Learning of Natural Language*. Springer-Verlag: Berlin.

David MW Powers, 1997a. Learning and application of differential grammars. *Computational Natural Language Learning* (CoNLL97).

David M W Powers, 1997b. Unsupervised learning of linguistic structure: an empirical evaluation. *International Journal of Corpus Linguistics* 2 (1), 91-131

David M.W. Powers. 1998. The Total Turing Test and the Loebner Prize. *Proceedings of the Joint Conferences on New Methods in Language Processing and Computational Natural Language Learning*, pages 279-280. https://aclweb.org/anthology/W98-1235.

David M.W. Powers, Richard Leibbrandt, Darius Pfitzner, Martin Luerssen, Trent Lewis, Arman Abrahamyan, Kate Stevens. 2008. Language teaching in a mixed reality games environment. *Proceedings of the 1st international conference on PErvasive Technologies Related to Assistive Environments*, pp 70-79. ACM.

Gerald Prince, 1982. *Narratology: The Forms and Functions of Narrative.* The Hague: Mouton, 1982.

Ehud Reiter and Robert Dale. 2000. *Building Natural Language Generation Systems*. Cambridge University Press, Cambridge, UK.

Mark Riedl, David Thue, & Vadim Bulitko, (2011). Game AI as Storytelling. In P. A. González-Calero & M. A. Gómez-Martín (Eds.), *Artificial Intelligence for Computer Games* (pp. 125-150). Springer: doi:10.1007/978-1-4419-8188-2_6

Marie-Laure Ryan, 2003, On Defining Narrative Media. *Image [&] Narrative*. Issue 6. Medium Theory.

Marie-Laure Ryan, 2007. Toward a definition of narrative. *The Cambridge companion to narrative*, pp.22-35.

Claude Sammut. 1985. Concept Development for Expert System Knowledge Bases, *The Australian Computer Journal,* Vol. 17(1): 49-55.

Huma Shah. 2011a. *Deception-detection and machine intelligence in Practical Turing tests*. PhD Thesis. University of Reading.

Huma Shah. 2011b. Turing's misunderstood imitation game and IBM's Watson success. Invited Keynote: 2[nd] *Towards a Comprehensive Intelligence Test* (TCIT) Symposium, AISB 2011 Convention, York, UK.

Huma Shah, Kevin Warwick, J. Vallverdú, and D. Wu. 2016. Can Machines Talk? Comparison of Eliza with Modern Dialogue Systems. *Computers in Human Behavior*, vol. 58 pages 278–295. http://dx.doi.org/10.1016/j.chb.2016.01.004

Luc Steels, 1998. The origins of syntax in visually grounded robotic agents. Artificial Intelligence 103(1-2):133-156.

Catherine J. Stevens, B. Pinchbeck, T. Lewis, M. Luerssen, D. Pfitzner, and D.M.W. Powers 2016. Mimicry and expressiveness of an ECA in human-agent interaction: familiarity breeds content! *Computational Cognitive Science* 2 (1):1. https://doi.org/10.1186/s40469-016-0008-2.

Andrea Tartaro and Justine Cassell. 2006. *Authorable Virtual Peers for Autism Spectrum Disorders.* Workshop on Language-Enabled Educational Technology at the 17th European Conference on Artificial Intelligence (ECAI06).

Andrea Tartaro and Justine Cassell. 2008. Playing with Virtual Peers: Bootstrapping Contingent Discourse in Children with Autism. *Proceedings of International Conference of the Learning Sciences*.

Alan M. Turing (1950) Computing Machinery and Intelligence. *Mind* 49:433-460.

Richard Wallace. 2008. The Anatomy of A.L.I.C.E. Chapter 13 in: R. Epstein, G. Roberts and G. Beber (Eds): *Parsing the Turing Test: Philosophical and Methodological Issues in the Quest for the Thinking Computer*. Springer.

Joseph Weizenbaum, 1966. ELIZA — A Computer Program for the Study of Natural Language Communication between Man and Machine. *CACM* 9: 36-45.

Terry Winograd. 1972. Procedures as a Representation for Data in a Computer Program for Understanding Natural Language*, Cognitive Psychology* Vol. 3 (1).

Jörg Wittwer, Matthias Nückles, and Alexander Renkl. 2010. Using a Diagnosis-Based Approach to Individualize Instructional Explanations in Computer-Mediated Communication. *Educational Psychology Review, 22*(1),: 9-23.

Dongqiang Yang and David M W Powers, 2008. Automatic thesaurus construction, *Australasian Conference on Computer Science.*